\documentclass[review]{elsarticle}

\usepackage{lineno,hyperref}
\modulolinenumbers[5]

\usepackage{graphicx}
\usepackage{url}

\usepackage[utf8]{inputenc} 
\usepackage[colorinlistoftodos]{todonotes}

\usepackage{epstopdf}

\usepackage{flushend}

\usepackage{geometry}

\usepackage{alltt}

\usepackage{amsmath,amssymb}
\usepackage{xfrac}

\usepackage{marginnote}
\usepackage{graphicx}

\usepackage{adjustbox}

\usepackage[boxruled,linesnumbered]{algorithm2e}

\usepackage{algorithmic}

\usepackage{url}
\usepackage{datetime}

\usepackage{afterpage}
\usepackage{wasysym}
\usepackage{hyperref}
\usepackage{xspace}

\usepackage{pifont}

\newtheorem{definition}{Definition}

\newcommand{\Probab}[1]{\mathrm{Pr}({#1})}
\newcommand{\Pcond}[2]{\Probab{{#1}\mid{#2}}}
\newcommand{\Pconj}[2]{\Probab{{#1} \wedge {#2}}}

\newcommand{\squishlist}{
	\begin{list}{$\bullet$} {
 		\setlength{\itemsep}{0pt}
     	\setlength{\parsep}{3pt}
     	\setlength{\topsep}{3pt}
     	\setlength{\partopsep}{0pt}
     	\setlength{\leftmargin}{2em}
     	\setlength{\labelwidth}{1.5em}
     	\setlength{\labelsep}{0.5em}
	}
}

\newcommand{\squishlisttight}{
	\begin{list}{$\bullet$} {
		\setlength{\itemsep}{0pt}
    	\setlength{\parsep}{0pt}
   		\setlength{\topsep}{0pt}
    	\setlength{\partopsep}{0pt}
    	\setlength{\leftmargin}{2em}
    	\setlength{\labelwidth}{1.5em}
    	\setlength{\labelsep}{0.5em}
	}
}

\newcommand{\squishdesc}{
	\begin{list}{} {
 		\setlength{\itemsep}{0pt}
     	\setlength{\parsep}{3pt}
     	\setlength{\topsep}{3pt}
     	\setlength{\partopsep}{0pt}
     	\setlength{\leftmargin}{1em}
		\setlength{\labelwidth}{1.5em}
		\setlength{\labelsep}{0.5em}
	}
}

\newcommand{\squishend}{
	\end{list}
}

\DeclareMathOperator*{\argmax}{arg\,max}

\newcommand{\DAG}{\textsc{dag}}

\newcommand{\NP}{\textit{NP}}

\newcommand{\bth}{\boldsymbol{\theta}}

\journal{Journal of Computational Science}

\begin{document}

\begin{frontmatter}


\title{Efficient computational strategies to learn the structure of probabilistic graphical models of cumulative phenomena}

\author[aff1]{Daniele Ramazzotti}
\author[aff2,aff3]{Marco S. Nobile}
\author[aff2,aff4]{Marco Antoniotti}
\author[aff2]{Alex Graudenzi}

\address[aff1]{Dept. of Pathology, Stanford University, CA, USA}
\address[aff2]{Dept. of Informatics, Systems and Communication, University of Milano - Bicocca, Milan, Italy}
\address[aff3]{SYSBIO Centre of Systems Biology, Milan, Italy}
\address[aff4]{Milan Center for Neuroscience, University of Milano - Bicocca, Monza, Italy}

\address{}



\begin{abstract}

Structural learning of Bayesian Networks (BNs) is a \emph{NP}-hard problem, which is further complicated by many theoretical issues, such as the $I$-equivalence among different structures.
In this work, we focus on a specific subclass of BNs, named Suppes-Bayes Causal Networks (SBCNs), which include specific structural constraints based on Suppes' probabilistic causation to efficiently model cumulative phenomena. 
\textcolor{black}{Here we compare} the performance, via extensive simulations, of various state-of-the-art search strategies, such as local search techniques and Genetic Algorithms, as well as of distinct regularization methods.  
The assessment is performed on a large number of simulated datasets from topologies with distinct levels of complexity, various sample size and different rates of errors in the data. 
Among the main results, we show that the introduction of Suppes' constraints dramatically improve the inference accuracy, by reducing  the solution space and providing a temporal ordering on the variables. We also report on trade-offs among different search techniques that can be efficiently employed in distinct experimental settings.
This manuscript is an extended version of the paper ``Structural Learning of Probabilistic Graphical Models of Cumulative Phenomena'' presented at the 2018 International Conference on Computational Science. \cite{ramazzotti2018structural}.

\end{abstract}




\end{frontmatter}




\section{Introduction}
\label{sect:introduction}

\emph{Bayesian Networks} (BNs) are probabilistic graphical models 
representing the relations of \emph{conditional dependence} among
random variables, encoded in \emph{directed acyclic graphs} (DAGs)
\cite{koller2009probabilistic}. In the last decades, BNs have been
effectively applied in several different fields and disciplines, such
as (but not limited to) diagnostics and predictive analytics
\cite{koller2009probabilistic}.


One of the most challenging task with BNs is that of \emph{learning}
their structure from data.  Two main approaches are commonly used to
tackle this problem.

\begin{enumerate}
\item \emph{Constraint-based} techniques: mainly due to the works by
  Judea Pearl \cite{pearl2009causality} and others, these approaches aim at
  discovering the relations of conditional independence from the data,
  using them as constraints to learn the network.

\item \emph{Score-based} techniques: in this case the problem of
  learning the structure of a BN is defined as an \emph{optimization}
  task (specifically, \emph{maximization}) where the search space of the
  valid solutions (i.e., all the possible DAGs) is evaluated via
  \emph{scores} based on a \emph{likelihood function}
  \cite{koller2009probabilistic}.
\end{enumerate}

Regardless of the approach, the main difficulty in this optimization
problem is the huge number of valid solutions in the search space,
namely, all the possible DAGs, which makes this task a known \emph{NP}-hard
one in its most general instance, and even when constraining each node
to have at most two parents
\cite{chickering1996learning,chickering2004large}. 
Therefore, all state-of-the-art techniques solve this task by means of
meta-heuristics
\cite{koller2009probabilistic,larranaga1996structure,teyssier2012ordering}.

Moreover, the inference is further complicated by the well-known issue of
\emph{$I$-equivalence}: BNs with even very different structures can
encode the same set of conditional independence properties\cite{koller2009probabilistic}.
Thus, any algorithm for structural
learning can converge to a set of equivalent structures rather than to
the correct one, given that the inference itself is performed by
learning the statistical relations among the variables emerging from
their induced distributions rather than the structure itself
\cite{koller2009probabilistic}.

In this paper, we investigate the application of BNs for the
characterization of a specific class of dynamical phenomena, i.e.,
those driven by the \emph{monotonic accumulation of events}. In
particular, the process being modeled/observed must imply:
\begin{enumerate}
\item a \emph{temporal ordering} among its events (i.e., the nodes in
  the BN), and

\item a monotonic accumulation over time, which probabilistically
  entails that the occurrence of an earlier event must be
  \emph{positively correlated} to the subsequent occurrence of its
  successors, leading to a \emph{significant temporal pattern}
  \cite{ramazzotti2016modeling}.
\end{enumerate}
An example can be found in the dynamics of \emph{cascading failures},
that is a failure in a system of interconnected parts where the
failure of a part can trigger the failure of successive parts.  These
phenomenon can happen in different contexts, such as power
transmission, computer networking, finance and biological systems.
In these scenarios, different configurations may lead to failure, but
some of them are more likely than others and, hence, can be modeled
probabilistically \cite{buldyrev2010catastrophic}. 

The two particular conditions mentioned above can be very well modelled by 
the notion of \emph{probabilistic causation} by Patrick Suppes 
\cite{suppes1970probabilistic,hitchcock2010probabilistic}, 
   and allow us to define a set of \emph{structural
  constraints} to the BNs to be inferred, which, accordingly, have
been dubbed as \emph{Suppes-Bayes Causal Networks} (SBCNs) in
previous works \cite{bonchi2015exposing,ramazzotti2016modeling}.
SBCNs have been already applied in a number of different fields,
ranging from cancer progression inference
\cite{loohuis2014inferring,ramazzotti2015capri,caravagna2016algorithmic}
to social discrimination discovery \cite{bonchi2015exposing} and stress testing \cite{gao2017efficient}. 

We specifically position our work within the aforementioned
optimization-based framework for BN structure learning. The goal of
this paper is to investigate how structure learning is influenced by
different algorithmic choices, when representing cumulative dynamical
phenomena.  In particular, it is known that given a \emph{temporal
  ordering} on the variables (i.e., a partially ordered set among the
events, \emph{poset} in the terminology of Bayesian networks) of a BN,
finding the optimal solution that is consistent with the ordering can
be accomplished in time $O(n^k)$, where $n$ is the number of variables
and $k$ the bounded in-degree of a node
\cite{buntine1991theory,cooper1992bayesian}.  Thus, the search in the
\emph{space of orderings} can be performed way more efficiently than
the search in the \emph{space of structures}, as the search space is
much smaller, the branching factor is lower and acyclicity checks are
not necessary \cite{teyssier2012ordering,ramazzotti2016parallel}. 

The determination of the right ordering 
  ordering.
   in complex dynamical phenomena is generally a difficult task, which often requires considerable domain knowledge.  
However, the representation of cumulative phenomena via SBCNs \textcolor{black}{allows one to overcome this hurdle}, as Suppes' constraints dramatically reduce the search space of valid solutions, also providing a temporal ordering on the variables.  
This represents a serious theoretical advancement in
structure learning of BNs for the modeling of cumulative phenomena,
which we investigate in this work with a series of synthetic
experiments.


In particular, in this paper we quantitatively assess the performance
of learning the structure of a BN when:
\begin{itemize}
\item the temporal ordering among variables is given / not given,
  i.e., when Suppes' constraints are imposed / not imposed (in the
  former case we deal with SBCNs);

\item different heuristic search strategies are adopted, i.e.,
  \emph{Hill Climbing} (HC), \emph{Tabu Search} (TS), and
  \emph{Genetic Algorithms} (GA);

\item different regularization terms are used, i.e., \emph{Bayesian
    Information Criterion} (BIC) and \emph{Akaike information
    criterion} (AIC).
\end{itemize}

\section{Background}
\label{sect:Background}

In this Section we provide an introduction to Bayesian networks together with a review of some state-of-the-art methods to tackle to problem of learning their structures from a set of \textcolor{black}{observations} $D$ over the variables described in the network.

\subsection{Bayesian Graphical Models}
\label{sec:bn_preliminaries}

A Bayesian network is a statistical graphical model that succinctly represents a \emph{joint distribution} over $n$ random variables and encodes it in a \textcolor{black}{\emph{directed acyclic graph}} $G = (V, E)$ over the $n$ nodes $V$ referring to the variables and their relations $E$ (arcs in the DAG).
Given the structure of a BN, the full joint distribution of the $n$ variables can be written as the product of the conditional distributions on each variable.  
In fact, an edge between pair of nodes, e.g., $A$ and $B$, denotes statistical dependence, i.e., $\Pconj{A}{B} \neq \Probab{A} \Probab{B}$, regardless of which any other variables we condition on, that is, for any other set of variables $\cal C$ it holds that \cite{koller2009probabilistic}
\textcolor{black}{\begin{equation}
  \Pcond{A \wedge B}{\mathcal{C}} \neq
  \Pcond{A}{\mathcal{C}} \cdot \Pcond{B}{\mathcal{C}}.
\end{equation}}

In such a DAG, the set of variables connected toward any node $X$ determines its set of ``parent'' nodes $\pi(X)$.
Moreover, the joint distribution over all the variables can be written
as $\prod_{X} \Pcond{X}{\pi(X)}$, where, if a node has no incoming edges (i.e., no parents), in the product we use its marginal probability $\Probab{X}$.  
Thus, to compute the probability of any combination of values over the variables, only  the conditional probabilities of each variable given its parents must be parameterized.  However, even in the simplest case of binary variables, the number of parameters in each conditional probability table is locally of exponential size: namely,\\$2^{|\pi(X)|} - 1$.
Thus, the total number of parameters needed to compute the full joint distribution is of size $\sum_{X} 2^{|\pi(X)|} - 1$, which is \textcolor{black}{considerably less} than $2^n-1$ for sparse networks.

A useful property of the graphical structure is that we can define, for each variable, a set of nodes called the \emph{Markov blanket} such that, conditioned on it, this variable is independent of all the other variables in the network. 
It can be proven that, for any BN, the Markov blanket consists of a node's parents, its children and the parents of the children \cite{koller2009probabilistic}.

We also point out that the usage of the symmetrical notion of conditional dependence introduces important limitations in the task of learning the structure of a BN. 
As a matter of fact, we note that the two edges $A \rightarrow B$ and $B \rightarrow A$ denote equivalent dependence between $A$ and $B$. 
Hence, two graphs having a different structure can model an identical  set of independence and conditional independence relations (\emph{I-equivalence}). 
This yields to the notion of \emph{Markov equivalence class} as a \emph{partially directed acyclic graph}, in which the edges that can take either orientation are left undirected.
It is also known that two BNs are Markov equivalent when they have the same \emph{skeleton} and the same \emph{$v$-structures}, the former being the set of edges, ignoring their direction (e.g., $A \to B$ and $B \to A$ constitute a unique edge in the skeleton) and the latter being all the edge structures in which a variable has at least two parents, but those do not share an edge (e.g., $A \rightarrow B \leftarrow C$) \cite{judea1991equivalence}.

BNs have an interesting relation to canonical boolean logical operators $\wedge$, $\vee$ and $\oplus$ and formulas over variables \cite{korsunsky2014inference,ramazzotti2016modeling}. 
In fact these formulas, which are ``deterministic'' in principle, in BNs are naturally softened into \emph{probabilistic relations} to allow some degree of uncertainty or noise. 
This probabilistic approach to modeling logic allows representation of qualitative relationships among variables in a way that is inherently robust to small perturbations by noise. 
For instance, the phrase \emph{``in order to hear music when listening to an mp3, it is necessary and sufficient that the power is on and the headphones are plugged in"} can be represented by a probabilistic conjunctive formulation that relates power, headphones and music, in which the probability that music is audible depends only on whether power and headphones are present. 
On the other hand, there is a small probability that the music will still not play (perhaps we forgot to load any songs into the device) even if both power and headphones are on, and there is small probability that we will hear music even without power or headphone (perhaps we are next to a concert and overhear that music)
\cite{korsunsky2014inference,ramazzotti2016model}.


\subsection{Approaches to Learning the Structure of a BN}
\label{sec:bn_learning}
In the the literature, there have been two initial families of methods aimed at learning the structure of a BN from data. 
The methods belonging to the first family aim to explicitly \emph{capture all the   conditional independence relations} encoded in the edges, and will be referred to as \emph{constraint based approaches} (\ref{sec:bn_learning_structural}). 
The second family, that of \emph{score based approaches} (\ref{sec:bn_learning_score}), aims at the selection of a model that \emph{maximizes the likelihood of the data} given the model. Since both  approaches lead to intractability (\NP{}-hardness)  \cite{chickering1996learning,chickering2004large}, computing and verifying an exact solution is impractical.
For this reason, heuristic methods like Hill Climbing \cite{skiena1998algorithm}, Tabu Search \cite{glover1989tabu}, Simulated Annealing \cite{kirkpatrick1984optimization} and Genetic Algorithms \cite{golberg1989genetic,john1992holland} are generally employed. 
These algorithms are characterized by a polynomial complexity, although they only provide asymptotic guarantees of converging to optimal solutions.

Recently, a third class of learning algorithms that takes advantage of \emph{specialized logical relations} (mentioned in the previous section) have been  introduced (\ref{sec:constrained_nets}). 
In the  rest of this section we describe in detail some of these approaches, leaving to specific readings more detailed discussions \cite{koller2009probabilistic,korsunsky2014inference,ramazzotti2016model}.

\subsubsection{Constraint-based Approaches}
\label{sec:bn_learning_structural}
We briefly present an intuitive explanation of several common algorithms used for structure discovery by explicitly considering conditional independence relations between variables. 
For more detailed explanations and analyses of complexity, correctness and stability, we refer the reader to the related references \cite{spirtes2000causation,tsamardinos2003algorithms}.

The basic idea behind this class of algorithms is to build a graph structure reflecting the independence relations in the observed data, thus matching as closely as possible the empirical distribution. 
The difficulty in this approach lies in the number of conditional pairwise independence tests that an algorithm would have to perform to test all possible relations.
This number is indeed \emph{exponential}, requiring to condition on a power set, when testing for the conditional independence between two variables.
Because of this inherent intractability, this class of algorithms requires the introduction of some \emph{approximations}.


\subsubsection{Score-based Approaches}
\label{sec:bn_learning_score}
These approaches to structural learning \textcolor{black}{aim to maximize the likelihood} of a set of observed data.
Since we assume that the data are independent and identically distributed, the likelihood of the data $\mathcal{L}(\cdot)$ is simply the product of the probability of each observation. 
That is,
\textcolor{black}{\begin{equation}
  \mathcal{L}(D) = \prod_{d\in D} \Probab{d}
\end{equation}}
for a set of observations $D$. Since we want to infer a model $\cal G$ that best explains the observed data, we define the likelihood of observing the data given a specific model $\cal G$ as:
\textcolor{black}{\begin{equation}
\mathcal{LL}(\mathcal{G},D) = \prod_{d \in D} \Pcond{d}{\mathcal{G}}\, .
\end{equation}}
However, the actual likelihood is never used in practice, as this quantity rapidly becomes very small and impossible to represent in a computer.
Instead, the logarithm of the likelihood function is usually adopted for three reasons: 
\emph{(i)} the $log(\cdot)$ function is monotonic; 
\emph{(ii)} log-likelihood mitigates the numerical issues caused by normal likelihood; 
\emph{(iii)} it is easy to compute, because the
logarithm of a product is equal to the sum of the logs (e.g., $\log(xy) =\log x + \log y $), and the likelihood for a Bayesian network is a product of simple terms \cite{koller2009probabilistic}.

Practically, however, there is a problem in learning the network structure by maximizing log-likelihood alone.  
Namely, for any arbitrary set of data, the most likely graph is always the fully connected one (i.e., all edges are present), since adding an edge can only increase the likelihood of the data.  
To overcome this limitation, log-likelihood is almost always supplemented with a
\emph{regularization term} that penalizes the complexity of the model. 
There is a plethora of regularization terms, some based on information theory and others on Bayesian statistics (see \cite{carvalho2009scoring} and references therein), which all serve to promote \emph{sparsity} in the learned graph structure, though different regularization terms are better suited for particular applications \cite{koller2009probabilistic,ramazzotti2016model}.

\subsubsection{Learning Logically Constrained Networks}
\label{sec:constrained_nets}
In Section \ref{sec:bn_preliminaries}, we noted that an important class of BNs captures common binary logical operators, such as $\wedge$, $\vee$, and $\oplus$. 
Although the learning algorithms mentioned above can be used to infer the structure of such networks, some algorithms employ knowledge of these logical constraints in the learning process.

A widespread approach for the learning of monotonic progression networks with a directed acyclic graph (DAG) structure and \emph{conjunctive events} are \emph{Conjunctive Bayesian Networks} (see CBNs, \cite{beerenwinkel2007conjunctive}). 
This approach was originally adopted to model cancer progression in terms of accumulation of drivers genes \cite{ramazzotti2015capri,caravagna2016algorithmic}, in a way closely related to the model we discuss in this work.


This model is a standard BN over Bernoulli random variables with the constraint that the probability of a node $X$ taking the value $1$ is zero if at least one of its parents has value $0$. 
This defines a conjunctive relationship, in that all the parents of $X$ must be $1$ for $X$ to possibly be $1$.  
Thus, this model alone cannot represent noise, which is an essential part of any real data. 
In response to this, \emph{hidden CBNs} \cite{gerstung2009quantifying} were developed by augmenting the set of variables: a correspondence to a new variable $Y$ that represents the observed state is assigned to each CBN variable $X$, which captures  the ``true" state. 
Thus, each new variable $Y$ takes the value of the corresponding variable $X$ with a high probability, and the opposite value with a low probability, in order to model noise observations. 
In this model, the variables $X$ are latent, i.e., they are not present in the observed data, and have to be inferred from the observed values for the new variables.





\section{Inference of Causal Networks}
\label{sect:Method}
In this Section we present the foundations of our framework and, specifically, we define the main characteristics of the SBCNs  and of some heuristic strategies for the likelihood fit. 
Without losing in generality, from now on, we consider a simplified formulation of the problem of learning the structure of BNs where all the variables depicted in the graph are Bernoulli random variables, i.e., their support is $(0,1)$. 
All the conclusions derived in these settings can be also directly applied to the general case where the nodes in the BN describe geneal random variables
\cite{koller2009probabilistic}.

More precisely, we consider as an input for our learning task a dataset $D$ of $n$ Bernoulli variables and $m$ cross-sectional samples. 
We assume the value $1$ to indicate that a given variable has been observed in the sample and $0$ that the variable had not been observed.


\subsection{Suppes-Bayes Causal Networks}
\label{sect:approach}
In \cite{suppes1970probabilistic}, Suppes introduced the notion of \emph{prima facie causation}. 
A prima facie relation between a cause $u$ and its effect $v$ is verified when the following two conditions are true.

\begin{enumerate}
\item \emph{Temporal Priority} (TP): any cause happens before its
  effect.
\item \emph{Probability Raising} (PR): the presence of the cause
  raises the probability of observing its effect.
\end{enumerate}

\begin{definition}[Probabilistic
  Causation,~\cite{suppes1970probabilistic}]
  \label{def:praising}
  For any two events $u$ and $v$, occurring respectively at times
  $t_u$ and $t_v$, under the mild assumptions that $0 < \Probab{u},
  \Probab{v} < 1$, the event $u$ is called a \emph{prima facie cause}
  of $v$ if it occurs \emph{before} and \emph{raises the probability}
  of $u$, i.e.,
  \begin{equation}
    \begin{cases}
      \mathrm{(TP)} \quad t_u < t_v \\
      \mathrm{(PR)} \quad \Pcond{v}{u} > \Pcond{v}{\neg u}
    \end{cases}
  \end{equation}
\end{definition}

\noindent
The notion of prima facie causality has known limitations in the
context of the general theories of causality
\cite{hitchcock2010probabilistic}, however, this characterization
seems to appropriate to model the dynamics of phenomena driven by the
monotonic accumulation of events where a temporal order among them is
implied and, thus, the occurrence of an early event positively
correlates to the subsequent occurrence in time of a later one.
Let us now refer again to systems where cascading failure may occur:
some configurations of events, in a specific order, may be more likely
to cause a failure than others.
This condition leads to the emergence of an observable \emph{temporal
  pattern} among the events captured by Suppes' definition of
causality in terms of statistical relevance, i.e., statistical
dependency.

Let us now consider a graphical representation of the aforementioned
dynamics in terms of a BN $G = (V,E)$.
Furthermore, let us consider a given node $v_i \in V$ and let us name
$\pi(v_i)$ the set of all the nodes in $V$ pointing to (and yet
temporally preceding) $v_i$. Then, the joint probability distribution
of the $n = \left\vert{V}\right\vert$ variables induced by the BN can
be written as:

\begin{equation}
  \Probab{\mathit{v_1, \ldots, v_n}}
  = \prod_{v_i \in V} \Probab{v_i | \pi(v_i)}
\end{equation}

When building our model, we need to constrain the characteristics of
the considered relations as depicted in the network (i.e., the arcs in
the graph), in order to account for the cumulative process above
mentioned, which, in turns, needs to be reflected in its induced
probability distribution \cite{ramazzotti2016modeling}. To this
extent, we can define a class of BNs over Bernoulli random variables
named \emph{Monotonic Progression Networks} (MPNs)
\cite{ramazzotti2016modeling,korsunsky2014inference,farahani2013learning}. Intuitively,
MPNs represent the progression of events monotonically\footnote{The
  events accumulate over time and when later events occur, earlier
  events are observed as well.} accumulating over time, where the
conditions for any event to happen is described by a probabilistic
version of the canonical boolean operators, i.e., conjunction
($\land$), inclusive disjunction ($\lor$), and exclusive disjunction
($\oplus$).

MPNs can model accumulative phenomena in a probabilistic fashion,
i.e., they are also modeling irregularities (noise) in the data as a
small probability $\varepsilon$ of not observing later events given
their predecessors.

%
%
%

Given these premises, in \cite{ramazzotti2015capri} the authors
describe an efficient algorithm (named CAPRI, see \cite{ramazzotti2015capri} 
for details) to learn the structure of constrained 
Bayesian networks which account for Suppes' criteria and which later
on are dubbed \emph{Suppes-Bayes Causal Networks} (SBCNs) in
\cite{bonchi2015exposing}. SBCNs are well suited to model cumulative
phenomena as they may encode irregularities in a similar way to MPNs
\cite{ramazzotti2016modeling}. The efficient inference schema of
\cite{ramazzotti2016modeling} relies on the observation (see
\cite{teyssier2012ordering}) that a way for circumventing the
intrinsic computational complexity of the task of learning the
structure of a Bayesian Network is to postulate a pre-determined
ordering among the nodes. Intuitively, CAPRI exploits Suppes' theory
to first mine an ordering among the nodes, reducing the complexity of
the problem, and then fits the network by means of likelihood
maximization.
In \cite{ramazzotti2016modeling} it is also shown that a SBCN, learned
using CAPRI, can also embed the notion of
accumulation through time as defined in a MPN, and, specifically,
conjunctive parent sets; nevertheless SBCNs can easily be generalized
to represent all the canonical boolean operators (\emph{Extended
  Suppes-Bayes Causal Networks}), notwithstanding an increase of the
algorithmic complexity \cite{ramazzotti2016modeling}.
We refer the reader to \cite{ramazzotti2016modeling} for further
details and, following \cite{bonchi2015exposing}, we now formally
define a SBCN.

\begin{definition}[Suppes-Bayes Causal Network]
  \label{def:scn}

  \noindent
  \emph{Given an input cross-sectional dataset $D$ of $n$ Bernoulli
    variables and $m$ samples, the Suppes-Bayes Causal Network $SBCN =
    (V,E)$ subsumed by $D$ is a directed acyclic graph such that the
    following requirements hold:}
  \begin{enumerate}
  \item \emph{\textbf{[Suppes' constraints]} for each arc $(u \to v)
      \in E$ involving the selective advantage relation between nodes
      $u,v \in V$, under the  mild assumptions that $0 < \Probab{u},
      \Probab{v} < 1$}:
    \begin{equation}
      \Probab{u} > \Probab{v} \quad \text{and} \quad \Pcond{v}{u} >
      \Pcond{v}{\neg u} \,.
    \end{equation}
  \item \emph{\textbf{[Simplification]} let $E'$ be the set of  arcs
      satisfying the Suppes' constraints as before; among all the
      subsets of $E'$, the set of arcs $E$ is the one whose
      corresponding graph maximizes the     \textcolor{black}{
log-likelihood $\mathcal{LL}$} of the data and the
      adopted regularization function $R(f)$:}
    \begin{equation}
      \label{eq:fitness}
      E = \argmax_{E \subseteq E', G =(V,E)}  \mathcal{LL}(G,D) - R(f) \,.
    \end{equation}
  \end{enumerate}
\end{definition}


Before moving on, we once again notice that the efficient
implementation of Suppes' constraints of CAPRI
 does not, in general, guarantee to converge to
the monotonic progression networks as depicted before. 
To overcome this
limitation, one could extend the Algorithm in 
order to learn, in addition to the network structure, also the logical
relations involving any parent set, increasing the overall
computational complexity.
Once again, we refer the interested reader to the discussions provided
in \cite{ramazzotti2015capri,ramazzotti2016modeling} and, without
losing in generality, for the purpose of this work, we consider the
efficient implementation of CAPRI presented in \cite{ramazzotti2015capri}. 

\textcolor{black}{It is important to remark that the evaluation of Suppes' constraints might be extended to longer serial dependence relations, by assessing, for instance, the statistical dependency involving more than two events. We here decide to evaluate pairwise conditions to keep the overall computational complexity at a minimum. However, we leave the investigation of this issue to further development of the framework}.

\subsection{Optimization and Evolutionary Computation}
\label{sec:opt-evolutionary}

The problem of the inference of SBCNs can be re-stated as an optimization problem, in which the goal is the maximization of a likelihood score.
Regardless of the strategy used in the inference process, the huge size of the search space of valid solutions makes this problem very hard to solve.
Moreover, as stated above, the \emph{general} problem of learning the structure of a BN is \NP{}-hard
\cite{chickering2004large}\footnote{We are aware of \emph{special}
  formulations of the problem that are solvable in polynomial time.  
  Their existence points to interesting questions regarding the ``barrier'' between  \NP{} problems and polynomial ones; however, these are questions beyond the scope of the present paper.}. 
Because of that, state-of-the-art techniques largely rely on heuristics \cite{koller2009probabilistic}, often based on stochastic global optimization methods like Genetic Algorithms (GAs) \cite{larranaga1996structure,ramazzotti2016modeling}.
Methods for BN learning can roughly be subdivided into two categories: single individual or population-based meta-heuristics.

Hill Climbing (HC) and Tabu Search (TS) both belong to the first category.
The former is a greedy approach for the structural learning of BNs, in which new edges are attached to the current putative solution as long as they increase the likelihood score and they do not introduce any cycles in the network.
TS is a stochastic variant of HC able to escape local minima, in which solutions visited in the past are not repeated by means of a tabu list.

GAs \cite{holland1975adaptation}, a global search methodology inspired by the mechanisms of natural selection, belong to the second category.
GAs were shown to be effective for BN learning, both in the case of available and not available \emph{a priori} knowledge about nodes' ordering \cite{larranaga1996structure,ramazzotti2016modeling}.
In a GA, a population $\mathfrak{P}$ of candidate solutions (named individuals) iteratively evolves, converging towards the global optimum of a given fitness function $f$ that, in this context, corresponds to the score to be
maximized.
%
The population $\mathfrak{P}$ is composed of $Q$ randomly created individuals, usually represented as fixed-length strings over a finite alphabet.
These strings encode putative solutions of the problem under investigation; in the case of BN learning, individuals represent linearized adjacency matrices of candidate BNs with $K$ nodes, encoded as string of binary values whose length\footnote{Since BNs are DAGs,  the representation can be reduced by not encoding the elements on  the diagonal, which are always equal to zero. In such case, the strings representing the individual have length $K \times K - K$.} is $O(K^2)$.
%

The individuals in $\mathfrak{P}$ undergo an iterative process whereby three genetic operators---selection, crossover and mutation---are applied in sequence to simulate the evolutionary recombination process, which results in a new population of possibly improved solutions. 
 During the selection process, individuals from $\mathfrak{P}$ are chosen, using a fitness-dependent sampling procedure \cite{back1994selective}, and are inserted into a new temporary population $\mathfrak{P}^{'}$.
In this work we assume a ranking selection: individuals are ranked according to their fitness values and the probability of selecting an individual is proportional to its \textit{position} in the ranking.
 The crossover operator is then used to recombine the structures of two promising selected parents.
We assume a single point crossover, in which the two strings encoded by the two parents are ``cut'' in the same random position and one of the resulting substrings is exchanged.
Finally, the mutation operator replaces an arbitrary symbol of an offspring, with a probability $\mathcal{P}_m$, using a random symbol taken from the alphabet.  
In the case of BNs, the mutation consists in flipping a single bit of the individual according to the specified probability.
It is worth noting that in the case of ordered nodes both  crossover and mutation are \emph{closed} operators, because the resulting offsprings always encode valid DAGs.  
To the aim of ensuring a consistent population of individuals throughout the generations, in the case of unordered nodes the two operators are followed by a correction procedure, in which the candidate BN is analyzed to identify the  presence of invalid cycles.  
For further information about our implementation of GAs for the inference of BNs, including the correction phase, we refer the interested reader to \cite{ramazzotti2016modeling}.

\section{Results}
\label{sect:Results}

We now discuss the results of a large number of experiments we
conducted on simulated data, with the aim of assessing the performance
of the state-of-the-art score-based techniques for the BN structure
inference, and comparing the performance of these methods with the
learning scheme defined in CAPRI.

Our main objective is to investigate how the performance is affected
by different algorithmic choices at the different steps of the
learning process.

\paragraph{Data Generation.} All simulations are performed with the
following generative models. We consider $6$ different topological
structures. 

\begin{enumerate}
\item \emph{Trees}: one predecessor at most for any node, one unique
  root (i.e., a node with no parents).
\item \emph{Forests}: likewise, more than one possible root.
\item \emph{Conjunctive DAGs with single root}: 3 predecessors at most
  for each node, all the confluences are ruled by logical
  conjunctions, one unique root.
\item \emph{Conjunctive DAGs with multiple roots}: likewise, possible
  multiple roots.
\item \emph{Disjunctive DAGs with single root}: 3 predecessors at most
  for each node, all the confluences are ruled by logical
  disjunctions, one unique root.
\item \emph{Disjunctive DAGs with multiple roots}: likewise, possible
  multiple roots.
\end{enumerate}

We constrain the induced distribution of each generative structure by
implying a cumulative model for either conjunctions or disjunctions,
i.e., any child node cannot occur if its parent set is not activated
as described for the MPN in the Method Section \ref{sect:Method}. For
each of these configurations, we generate $100$ random
structures. Furthermore, we simulate a model of noise in terms of
random observations (i.e., false positives and false negatives)
included in the generated datasets with different rates. 

These data generation configurations are chosen to reflect: $(a)$
different structural complexities of the models in terms of number of
parameters, i.e., arcs, to be learned, $(b)$ different types of
induced distributions suitable to model cumulative phenomena as
defined by the MPNs (see Section \ref{sect:approach}), i.e.,
conjunction ($\land$) or inclusive disjunction ($\lor$)\footnote{Here
  we stick with the efficient search scheme of CAPRI
   and, for this reason, we do not consider
  exclusive disjunction ($\oplus$) parent sets} and, $(c)$ situations
of reduced sample sizes and noisy data. 

We here provide an example of data generation. Let now $n$ be the
number of nodes we want to include in the network and let us set
$p_{\min}=0.05$ and $p_{\max}=0.95$ as the minimum and maximum
probabilities of any node. A {\em directed acyclic graph without
  disconnected components} (i.e., an instance of types $(3)$ and $(5)$
topologies) with maximum depth $\log n$ and where each node has at
most $w^\ast = 3$ parents is generated. 

\begin{algorithm}[!ht]
\caption{Data generation: single source directed acyclic graphs}
\label{algo:data_generation_pseudo}
\KwIn{$n$, the number of nodes of the graph, $p_{\min}=0.05$ and
  $p_{\max}=0.95$ be the minimum and maximum probabilities of any node
  and $w^\ast = 3$ the maximum incoming edges per node.}

\KwResult{a randomly generated single source directed acyclic graph.}
 Pick an event $r\in G$ as the root of the directed acyclic graph\;
 Assign to each node $u \neq r$ an integer in the interval $[2, \lceil
 {\log n} \rceil]$ representing its depth in the graph  ($1$ is
 reserved for $r$), ensuring that each level has at least one node\;

\ForAll{nodes $u \neq r$}{ %
Let $l$ be the level assigned to the node\;
Pick $|\Probab{u}|$  uniformly  over $(0,w^\ast]$, and accordingly
define the parents of $u$ with events selected among those at which
level $l-1$ was assigned\;}
Assign $\Probab{r}$, a random value in the interval
$[p_{\min},p_{\max}]$\;
\ForAll{events $u \neq r$}{ %
Let $\alpha$ be a random value in the interval $[p_{\min},p_{\max}]$\; 
Let $\pi(u)$ be the direct predecessor of $u$\;
Then assign:
\[
\Probab{u} =  \alpha \Probab{x \in \pi(u)}\, ;
\]}
\KwRet{The generated single source directed acyclic graph.}
\end{algorithm}

\paragraph{Performance Assessment.} In all these configurations, the
performance is assessed in terms of:

\begin{itemize}
\item accuracy = $\frac{(TP + TN)}{(TP + TN + FP + FN)}$;
\item sensitivity = $\frac{TP}{(TP + FN)}$;
\item specificity = $\frac{TN}{(FP + TN)}$;
\end{itemize}
with $TP$ and $FP$ being the true and false positives (we mark as
positive any arc that is present in the network) and $TN$ and $FN$
being the true and false negatives (we mark negative any arc that is
not present in the network) with respect to the generative model. All
these measures are values in $[0,1]$ with results close to $1$
indicators of good performance.

\paragraph{Implementation.} In all the following experiments, the
adopted likelihood functions (i.e., the fitness evaluations) are
implemented using the \emph{bnlearn} package \cite{scutari2009learning}
written in the R language, while GA \cite{holland1975adaptation}
the \emph{inspyred} \cite{garrett2012inspyred}, \emph{networkx}
\cite{schult2008exploring} and \emph{numpy} \cite{oliphant2006guide}
packages.

\textcolor{black}{The framework for the inference of SBCNs is implemented in R and is available in the TRONCO suite for TRanslational ONCOlogy \cite{de2016tronco,antoniotti2016design}. TRONCO is available under a GPL3 license at its webpage: \href{https://sites.google.com/site/troncopackage}{https://sites.google.com/site/troncopackage} or on Bioconductor.}

\paragraph{Algorithm Settings.} We test the performance of classical
search strategies, such as Hill Climbing (HC) and Tabu Search (TS),
and of more sophisticated algorithms such Genetic Algorithms
(GA)\footnote{Further experiments on multi-objective optimization
  techniques, such as \emph{Non-dominated Sorting Genetic Algorithm}
  (NSGA- II), were performed, but are not shown here because of the
  worse overall performance, and of the higher computational cost,
  with respect to canonical GA.}.

For HC and TS, we generate data as described above with networks of
$10$ and $15$ nodes (i.e., $0/1$ Bernoulli random variables). We
generated $10$ independent datasets for each combination of the $4$
sample levels (i.e., $50$, $100$, $150$ and $200$ samples) and the $9$
noise levels (i.e., from $0\%$ to $20\%$ with step $2.5\%$) for a
total of $4,320,000$ independent datasets. The experiments were
repeated either $(i)$ including or $(ii)$ not including the Suppes'
constraints described in CAPRI \cite{ramazzotti2015capri}, and
independently using $5$ distinct optimization scores and
regularizators, namely standard $(i)$ log-likelihood
\cite{koller2009probabilistic}, $(ii)$ AIC
\cite{akaike1998information}, $(iii)$ BIC
\cite{schwarz1978estimating}, $(iv)$ BDE \cite{heckerman1995learning}
and $(v)$ K2 \cite{cooper1991bayesian}, leading to a final number of
$86,400,000$ different configurations. 

\textcolor{black}{Being more precise, given an input dataset of observations $D$ and a graphical model $G$, we can define a function to evaluate the goodness of this structure given the data:
\textcolor{black}{\begin{equation}
f(G, D) = \mathcal{LL}(D|G) - \mathcal{R}(G),
\end{equation}}
where $\mathcal{LL}(\cdot)$ is the log-likelihood, while $\mathcal{R}(\cdot)$ is a regularization~term with the aim of limiting the complexity of $G$.
The \DAG\ induced by $G$ in fact defines a probability distribution over its nodes, namely $\{x_1, \ldots, x_n\}$:
\textcolor{black}{\begin{equation}
\Probab{x_1, \ldots, x_n} = \prod_{x_i=1}^n \Pcond{x_i}{\pi_i},
\end{equation}\begin{equation}
\Pcond{x_i}{\pi_i} = \bth_{x_i\mid \pi_i},
\end{equation}}
where $\pi_i = \{ x_j \mid x_j \to x_i \in G\}$ are $x_i$'s parents in the DAG, and $\bth_{x_i\mid \pi(x_i)}$ is a density function. Then, the log-likelihood of the graph can be defined as:
\textcolor{black}{\begin{equation}
LL(D|G) = log \Pcond{D}{G,\bth} \, .
\end{equation}}
Then, the regularization term $\mathcal{R}(G)$ introduces a penalty for the number of parameters in the model $G$ also considering the size of the data. The above mentioned scores that we considered differ in this penalty, with AIC and BIC being Information-theoretic score and BDE and K2, Bayesian scores \cite{carvalho2009scoring}.}

While a detailed description of
these regularizators is beyond the scope of this paper, we critically
discuss the different performances granted by each strategy for the
inference of BNs. 

With respect to GA we used a restricted data generation settings,
using networks of $15$ nodes, datasets of $100$ samples and $5$ noise
levels (from $0\%$ to $20\%$ with step $5\%$) for a total of $3,000$
independent datasets. We tested the GA either $(i)$ with or $(ii)$
without Suppes' constraints, using BIC regularization term,  leading
to the final total of $6,000$ different configurations.
Finally, the GA was launched with a population size of $32$
individuals, a mutation rate of $p_m=0.01$ and $100$ generations.

\begin{figure*}[!ht]
\centering
\includegraphics[width=1.00\textwidth]{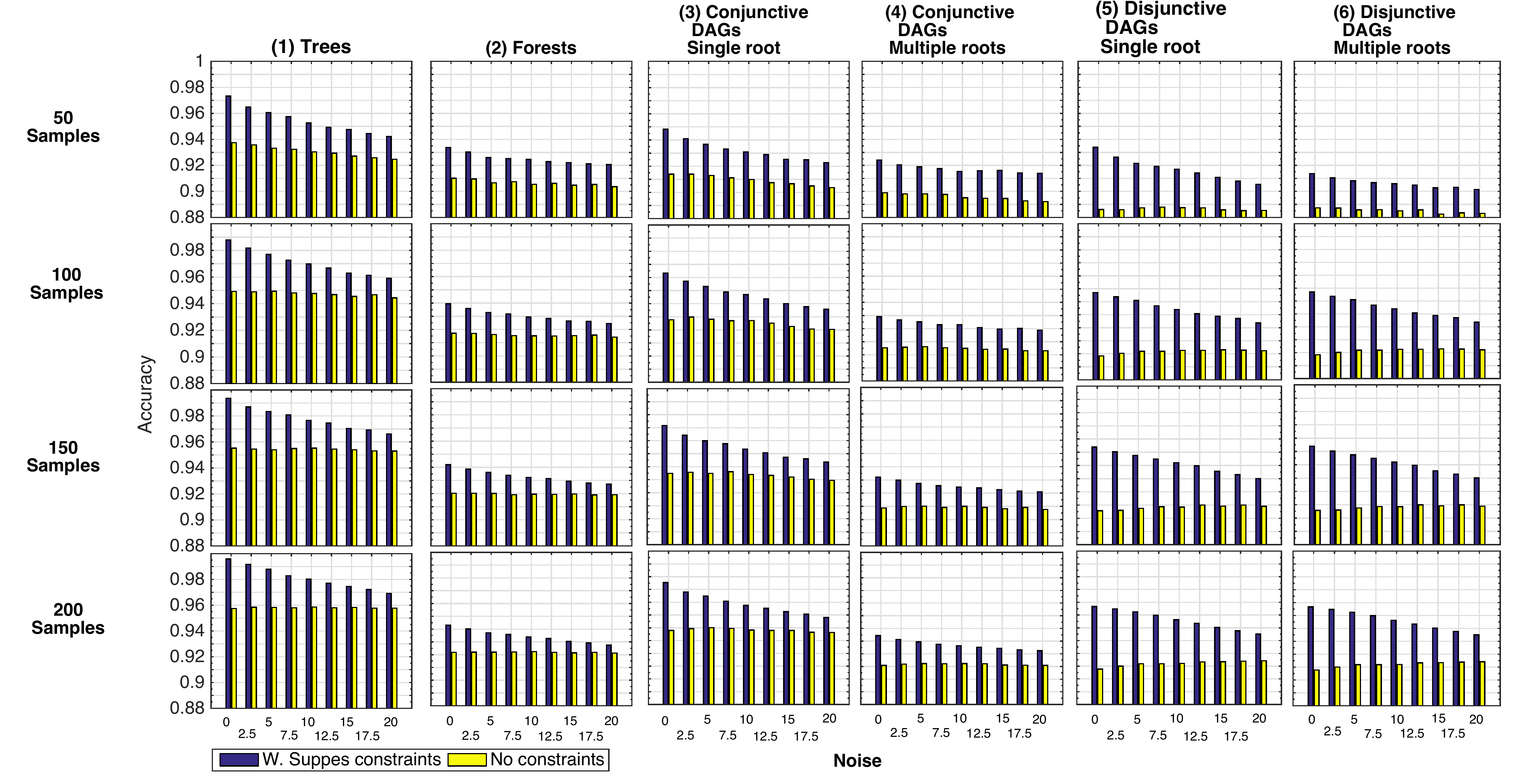}
\caption{Performance in terms of accuracy for the $6$ considered
  structures with $15$ nodes, noise levels from $0\%$ to $20\%$ with
  step $2.5\%$ and sample sizes of $50$, $100$, $150$ and $200$
  samples. Here we use BIC as a regularization scheme and we consider
  HC as a search stategy both for the classical case and when Suppes'
  priors are applied.}
\label{fig:results_figure_1}
\end{figure*}

\begin{figure*}[!ht]
\centering
\includegraphics[width=1.00\textwidth]{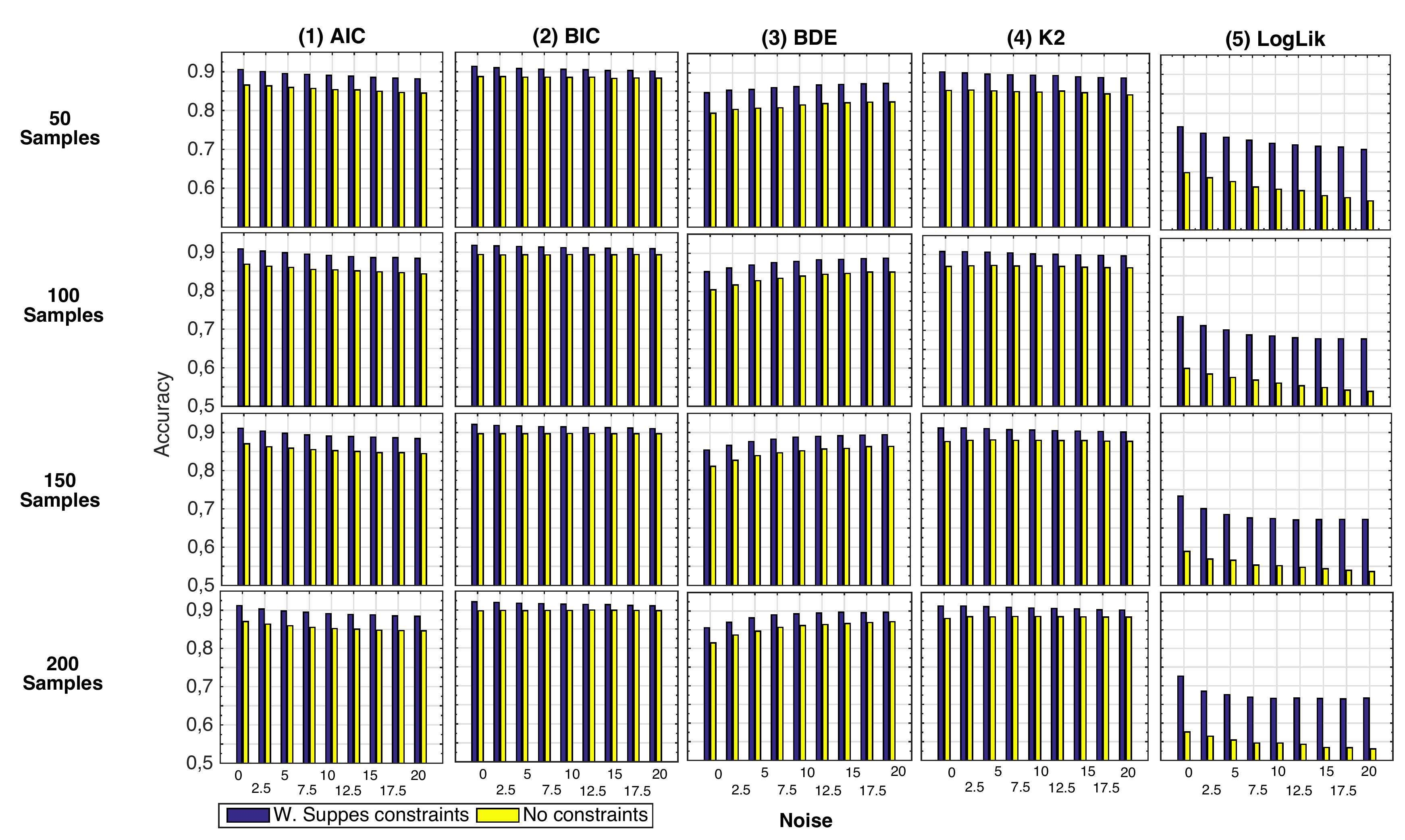}
\caption{Performance in terms of accuracy for directed acyclic graphs
  with multiple sources and disjunctive parents (structure $vi$) of
  $15$ nodes, noise levels from $0\%$ to $20\%$ with step $2.5\%$ and
  sample sizes of $50$, $100$, $150$ and $200$ samples. Here we
  consider HC as a search strategy both for the classical case and
  when Suppes' priors are applied and \textcolor{black}{we show the results for all five regularizators introduced in the text}.}
\label{fig:results_figure_2}
\end{figure*}

\begin{figure}[!ht]
\centering
\includegraphics[width=0.90\textwidth]{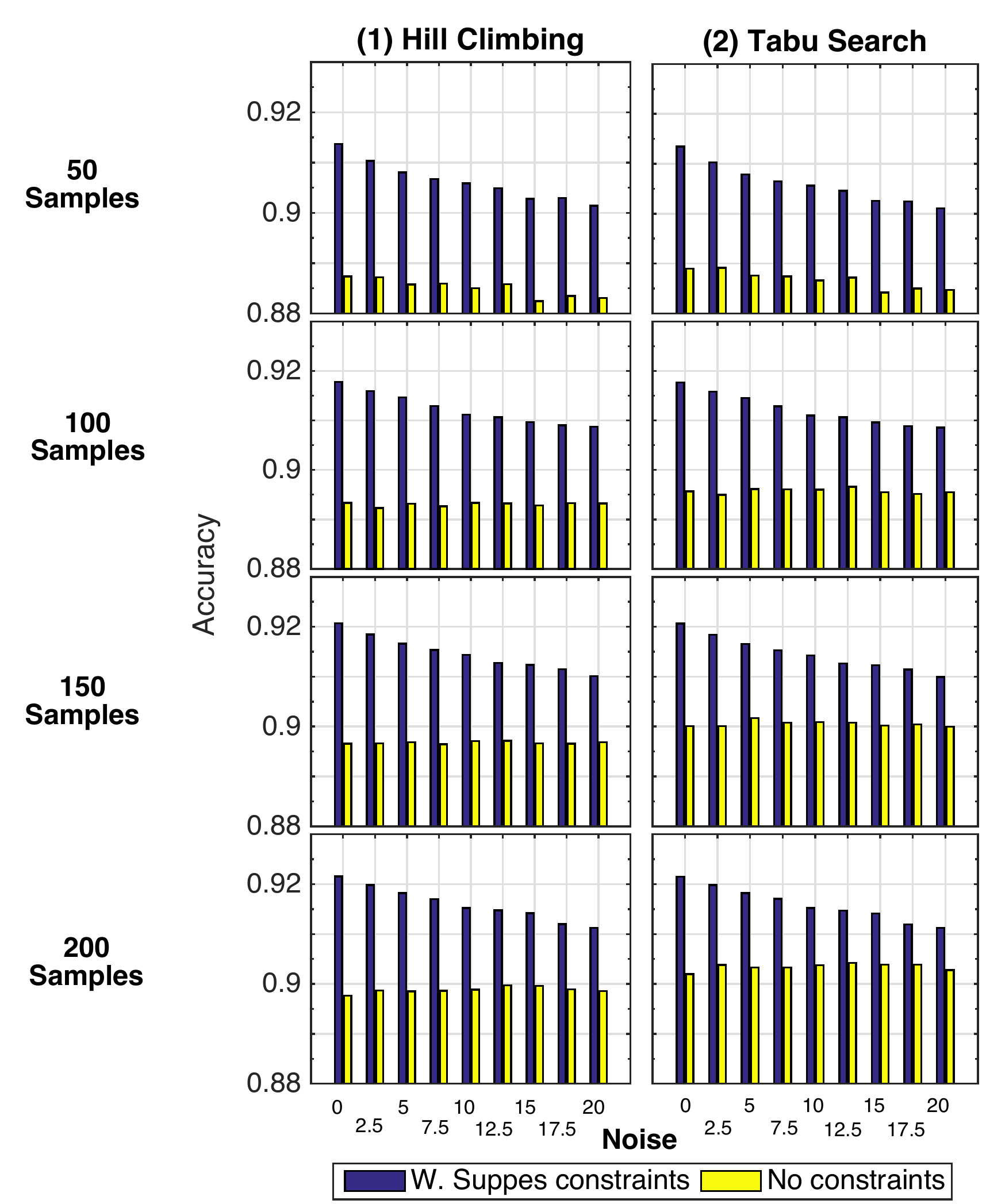}
\caption{Performance in terms of accuracy for directed acyclic graphs
  with multiple sources and disjunctive parents (structure $vi$) of
  $15$ nodes, noise levels from $0\%$ to $20\%$ with step $2.5\%$ and
  sample sizes of $50$, $100$, $150$ and $200$ samples. Here we use
  BIC as a regularization scheme and we  consider both HC and TB as
  search strategies for the classical case and when Suppes' priors are
  applied.}
\label{fig:results_figure_3}
\end{figure}

\begin{figure*}[!ht]
\centering
\includegraphics[width=1.00\textwidth]{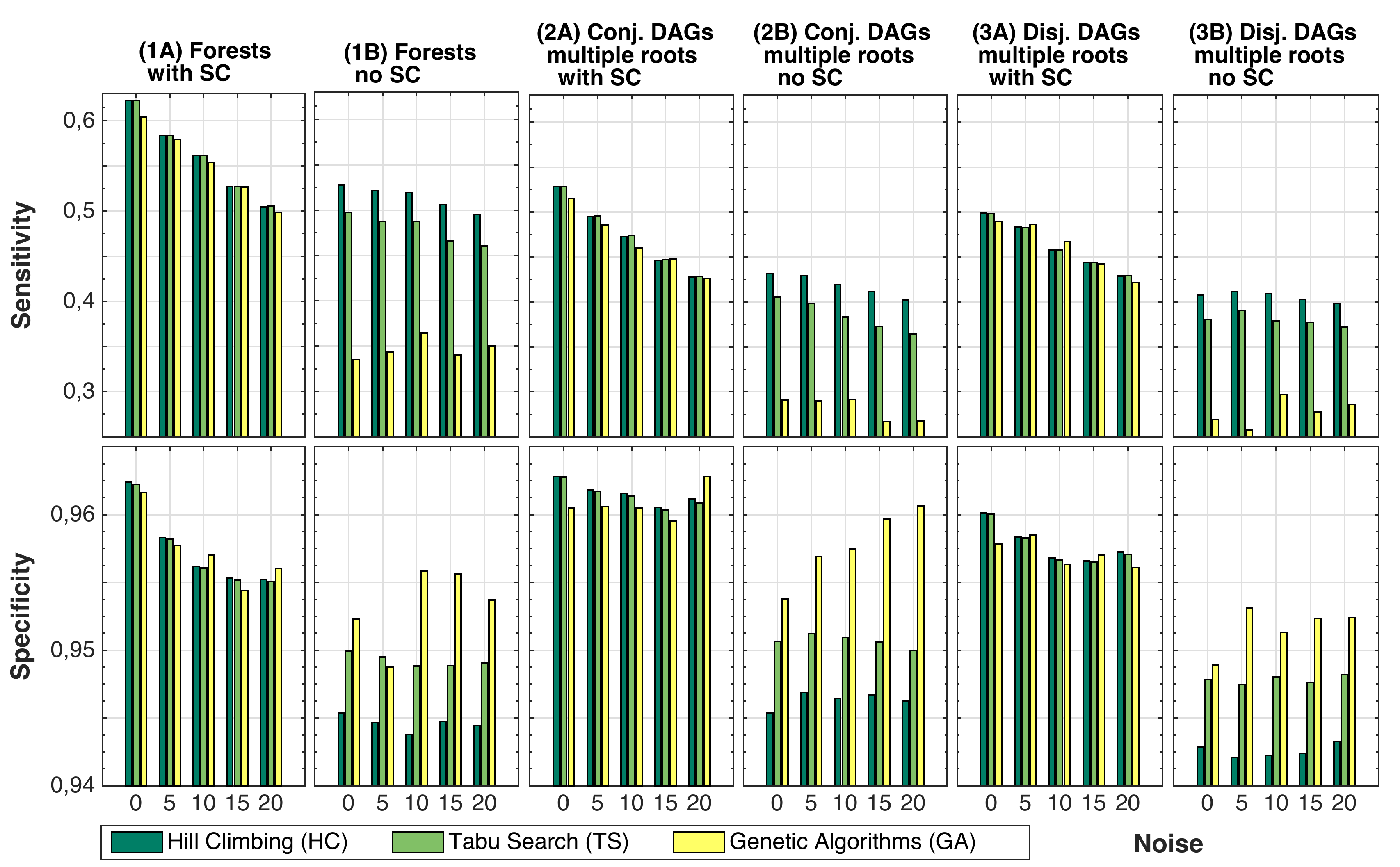}
\caption{Performance of HC, TB and GA, in terms of sensitivity and
  specificity for forests (panels 1A, 1B), directed acyclic graphs
  with multiple sources and conjunctive parents (panels 2A, 2B) and
  directed acyclic graphs with multiple sources and disjunctive
  parents (panels 3A, 3B) (configurations $(ii)$, $(iv)$ and $(vi)$)
  of $15$ nodes, noise levels from $0\%$ to $20\%$ with step $5\%$ and
  sample sizes of $100$ samples. Here we use BIC as a regularization
  scheme for HC and TB, and for all the algorithms we either consider
  (panels A) or not consider (panels B) Suppes' constraints (SC).}
\label{fig:results_figure_4}
\end{figure*}

We summarize the performance evaluation of the distinct techniques and
settings in the next Subsections and in Figures
\ref{fig:results_figure_1}, \ref{fig:results_figure_2},
\ref{fig:results_figure_3} and \ref{fig:results_figure_4}.

\subsection*{Performance Assessment}
\label{sec:performance-assessment}

By looking at Figure \ref{fig:results_figure_1}, one can first
appreciate the variation of accuracy with respect to a specific search
strategy, i.e., HC with BIC, which is taken as an example of typical
behavior.
In brief, the overall performance worsens with respect to: $(i)$ a
larger number of nodes in the network, $(i)$ more complex generative
structures, and $(iii)$ smaller samples sizes / higher noise rates. 
Although such a trend is intuitively expected, given the larger number
of parameters to be learned for more complex models, we here underline
the role of statistical complications, such as the presence of
\emph{spurious correlations} \cite{pearson1896mathematical} and the
occurrence of \emph{Simpson's paradox} \cite{good1987amalgamation}.

For instance, it is interesting to observe a typical decrease of the
accuracy when we compare topologies with the same properties, but
different number of roots (i.e., $1$ root vs. multiple roots). In the
former case, we expect, in fact, a lower number of arcs (i.e.,
dependencies) to be learned (on average) and, hence, we may attribute
the decrease of the performance to the emergence of spurious
correlations among independent nodes, such as the children of the
different sources of the DAG.  This is due to the fact that, when
sample sizes are not infinite, it is very unlikely to observe perfect
independence and, accordingly, the likelihood scores may lead to
overfitting.  The trends displayed in Figure
\ref{fig:results_figure_1} are shared by most of the analyzed search
strategies.

\paragraph{Role of the Regularization Term.} By looking at Figure
\ref{fig:results_figure_2} one can first notice that the accuracy with
no regularization is dramatically lower than the other cases, as a
consequence of the expected overfitting (in this case we compare the
performance of HC on disjunctive DAGs with multiple roots, but the
trend is maintained in the other cases).  Conversely, all
regularization terms ensure the inference of sparser models, by
penalizing the number of retrieved arcs.
BDE regularization term seems to be the only exception (see Figure
\ref{fig:results_figure_2}), leading to unintuitive behaviors: in
fact, while for all the other methods the performance decreases when
higher level of noise are applied, for BDE the accuracy seems to
improve with higher noise rates. 
This result might be explained by observing that given a topological
structure, structural spurious correlations may arise between a given
node and any of its undirected predecessors (i.e., one of the
predecessors of its direct parents): with higher error rates, and,
accordingly, more random samples in the datasets, all the correlations
are reduced, hence leading to a lower impact of the regularization
term.
Given these considerations, one can hypothesize that the overall trend
of BDE is due to a scarce penalization to the likelihood fit, favoring
dense networks rather than sparse ones.

\paragraph{Search Strategies.} No significant differences in the
performance between the accuracy of HC, TS and GA are observed.
However, one can observe a consistent improvement in sensitivity when
using GA (see Figures \ref{fig:results_figure_3} and
\ref{fig:results_figure_4}). This suggests different inherent
properties of the search schemes: while  with HC and TB the
regularization terms, rather than the search strategy, account for
most of the inference performance,  GAs are capable of returning
denser networks with better hit rates.  This is probably due to GA's
random mutations, which allow jumps into areas of the search space
characterized by excellent fitness, which could not be reached by
means of greedy approaches like HC.


\paragraph{Suppes' Structural Constraints.} Finally, the most
important result, which can be observed across all the different
experiments, is that the overall performance of all the considered
search strategies is dramatically enhanced by the introduction of
Suppes' structural constraints.
In particular, as one can see, e.g., in Figure
\ref{fig:results_figure_1}, there is a constant improvement in the
inference, up to $10\%$, when Suppes' priors are used. 
Even though the accuracy of the inference is affected by the noise in
the observations, in fact, the results with Suppes' priors are
consistently better than the inference with no constraints, with
respect to all the considered inference settings and to all the
performance measures. 
This is an extremely important result as it proves that the
introduction of structural constraints based on Suppes' probabilistic
causation indeed simplify the optimization task, by reducing the huge
search space, when dealing with BNs describing cumulative phenomena.

\section{Conclusion}
\label{sect:Conclusion}

In this paper we investigated the
structure learning of Bayesian Networks aimed at modeling phenomena
driven by the monotonic accumulation of events over time. To this end,
we made use of a subclass of constrained Bayesian networks named
Suppes-Bayes Causal Networks, which include structural constraints
grounded in Suppes' theory of probabilistic causation.

While the problem of learning the structure of a Bayesian Network is
known to be intractable, such constraints allow to prune the search
space of the possible solutions, leading to a tremendous reduction of
the number of valid networks to be considered, hence taming the
complexity of the problem in a remarkable way. 

We here discussed the theoretical implications of the inference
process at the different steps, also by comparing various
state-of-the-art algorithmic approaches and regularization methods.
We finally provided an in-depth study on realistically simulated data
of the effect of each inference choice, thus providing some sound
guidelines for the design of efficient algorithms for the inference of
models of cumulative phenomena. 

\textcolor{black}{According to our results, none of the tested search strategies significantly outperforms the others in all the experimental settings, in terms of both sensitivity and specificity.} 


Yet, we could prove that Suppes' constraints
consistently improve the inference accuracy, in all the considered
scenarios and with all the inference schemes, hence positioning SBCNs
as the new benchmark in the the efficient inference and representation
of cumulative phenomena. 



\subsubsection*{Acknowledgments.} This work was supported in part by the ASTIL Program of
    Regione Lombardia, by the ELIXIR-ITA network, and by the SysBioNet project, a MIUR initiative for the Italian Roadmap of European Strategy Forum on Research Infrastructures (ESFRI). We would like to thank for the useful discussions our colleagues 
  Giulio Caravagna of ICR, London, UK, Giancarlo Mauri of DISCo, 
  Universit\`{a} degli Studi di Milano-Bicocca, Milan, Italy, and Bud 
  Mishra of Courant Institute of Mathematical Sciences, New York 
  University, NY, USA. 

\clearpage
\label{sect:bib}
\section*{References}

\bibliography{References}

\end{document}